\documentclass[conference]{IEEEtran}
\IEEEoverridecommandlockouts

\usepackage{cite}
\usepackage{amsmath,amssymb,amsfonts}
\usepackage{algorithmic}
\usepackage{graphicx}
\usepackage{textcomp}
\usepackage{xcolor}
\usepackage{subcaption}
\usepackage{hyperref} 

\def\BibTeX{{\rm B\kern-.05em{\sc i\kern-.025em b}\kern-.08em
    T\kern-.1667em\lower.7ex\hbox{E}\kern-.125emX}}

\begin{document}

\title{A4-Unet: Deformable Multi-Scale Attention Network for Brain Tumor Segmentation}

\author{
\IEEEauthorblockN{Ruoxin Wang\textsuperscript{†}}
\IEEEauthorblockA{\textit{BNU-HKBU United International College}\\
Zhuhai, China \\
ruoxinwaaang@gmail.com}
\and
\IEEEauthorblockN{Tianyi Tang\textsuperscript{†}}
\IEEEauthorblockA{\textit{BNU-HKBU United International College}\\
Zhuhai, China \\
trumantytang@163.com}
\and
\IEEEauthorblockN{Haiming Du}
\IEEEauthorblockA{\textit{BNU-HKBU United International College}\\
Zhuhai, China \\
jennyduuu@163.com}
\and
\IEEEauthorblockN{Yuxuan Cheng}
\IEEEauthorblockA{\textit{BNU-HKBU United International College}\\
Zhuhai, China \\
t330201601@mail.uic.edu.cn}
\and
\IEEEauthorblockN{Yu Wang}
\IEEEauthorblockA{\textit{Sun Yat-sen Memorial Hospital}\\
Guangzhou, China \\
wangy2298@mail2.sysu.edu.cn}
\and
\IEEEauthorblockN{Lingjie Yang}
\IEEEauthorblockA{\textit{Sun Yat-sen Memorial Hospital}\\
Guangzhou, China \\
yanglj53@mail2.sysu.edu.cn}
\and
\IEEEauthorblockN{Xiaohui Duan}
\IEEEauthorblockA{\textit{Sun Yat-sen Memorial Hospital}\\
Guangzhou, China \\
yanglj53@mail2.sysu.edu.cn}
\and
\IEEEauthorblockN{Yunfang Yu}
\IEEEauthorblockA{\textit{Sun Yat-sen Memorial Hospital}\\
Guangzhou, China \\
yanglj53@mail2.sysu.edu.cn}
\and
\IEEEauthorblockN{Yu Zhou}
\IEEEauthorblockA{\textit{Shenzhen University}\\
Shenzhen, China \\
yu.zhou@szu.edu.cn}
\and
\IEEEauthorblockN{Donglong Chen\textsuperscript{*}}
\IEEEauthorblockA{\textit{BNU-HKBU United International College}\\
Zhuhai, China \\
donglongchen@uic.edu.cn}
\thanks{\textsuperscript{†}These authors contributed equally to this work.}
\thanks{\textsuperscript{*}Corresponding author.}
}

\maketitle

\begin{abstract}
Brain tumor segmentation models have aided diagnosis in recent years. However, they face MRI complexity and variability challenges, including irregular shapes and unclear boundaries, leading to noise, misclassification, and incomplete segmentation, thereby limiting accuracy. To address these issues, we adhere to an outstanding Convolutional Neural Networks (CNNs) design paradigm and propose a novel network named \textbf{A4-Unet}. In A4-Unet, Deformable Large Kernel Attention (DLKA) is incorporated in the encoder, allowing for improved capture of multi-scale tumors. Swin Spatial Pyramid Pooling (SSPP) with cross-channel attention is employed in a bottleneck further to study long-distance dependencies within images and channel relationships. To enhance accuracy, a Combined Attention Module (CAM) with Discrete Cosine Transform (DCT) orthogonality for channel weighting and convolutional element-wise multiplication is introduced for spatial weighting in the decoder. Attention gates (AG) are added in the skip connection to highlight the foreground while suppressing irrelevant background information. The proposed network is evaluated on three authoritative MRI brain tumor benchmarks and a proprietary dataset, and it achieves a 94.4\% Dice score on the BraTS 2020 dataset, thereby establishing multiple new state-of-the-art benchmarks. The code is available here: \href{https://github.com/WendyWAAAAANG/A4-Unet}{https://github.com/WendyWAAAAANG/A4-Unet}.
\end{abstract}


\begin{figure}[!ht]
	\begin{minipage}{0.32\linewidth}
		\centerline{\includegraphics[width=\textwidth]{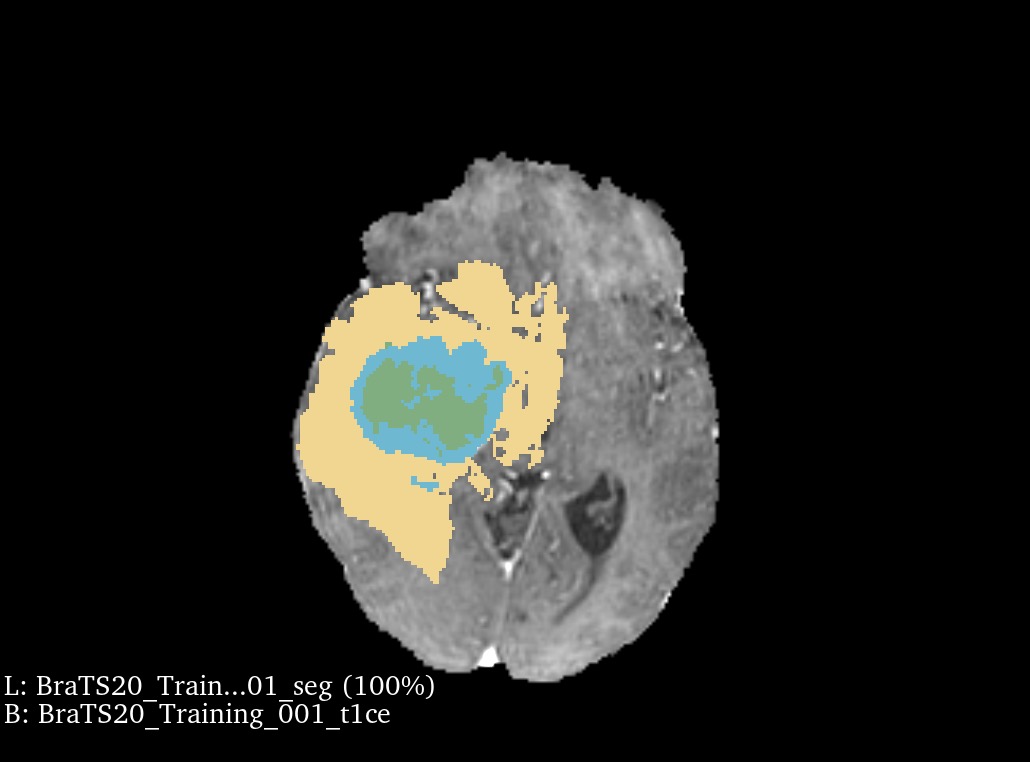}}
        \label{fig:mri}
	\end{minipage}
	\begin{minipage}{0.32\linewidth}
		\centerline{\includegraphics[width=\textwidth]{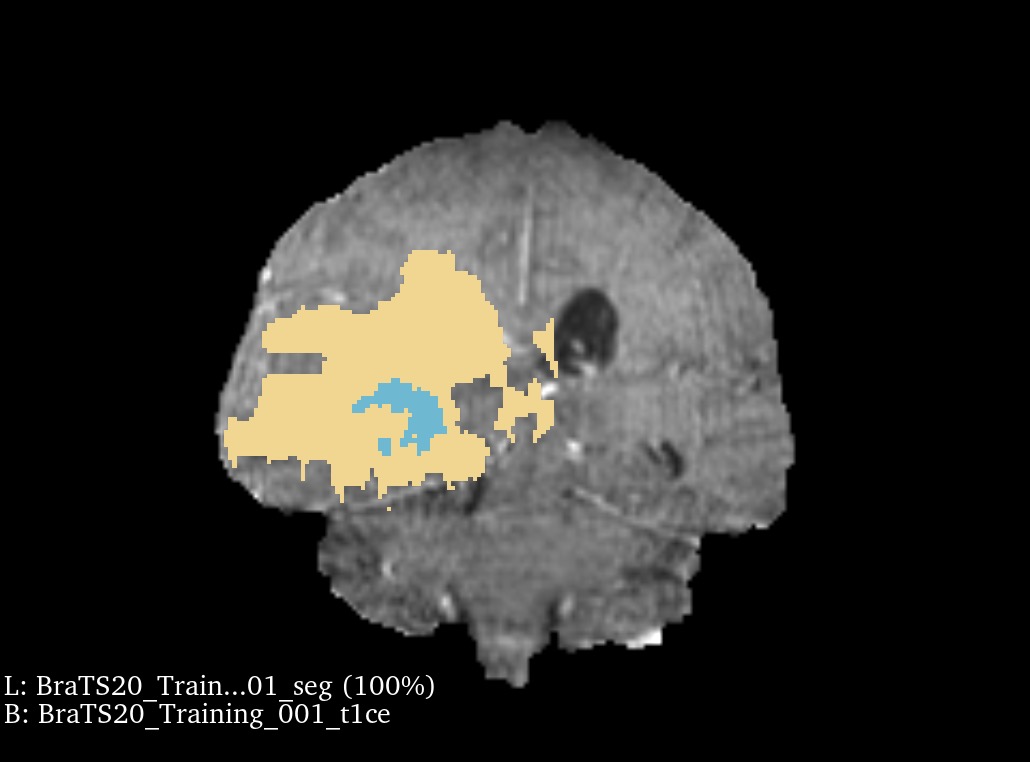}}
	\end{minipage}
	\begin{minipage}{0.32\linewidth}
		\centerline{\includegraphics[width=\textwidth]{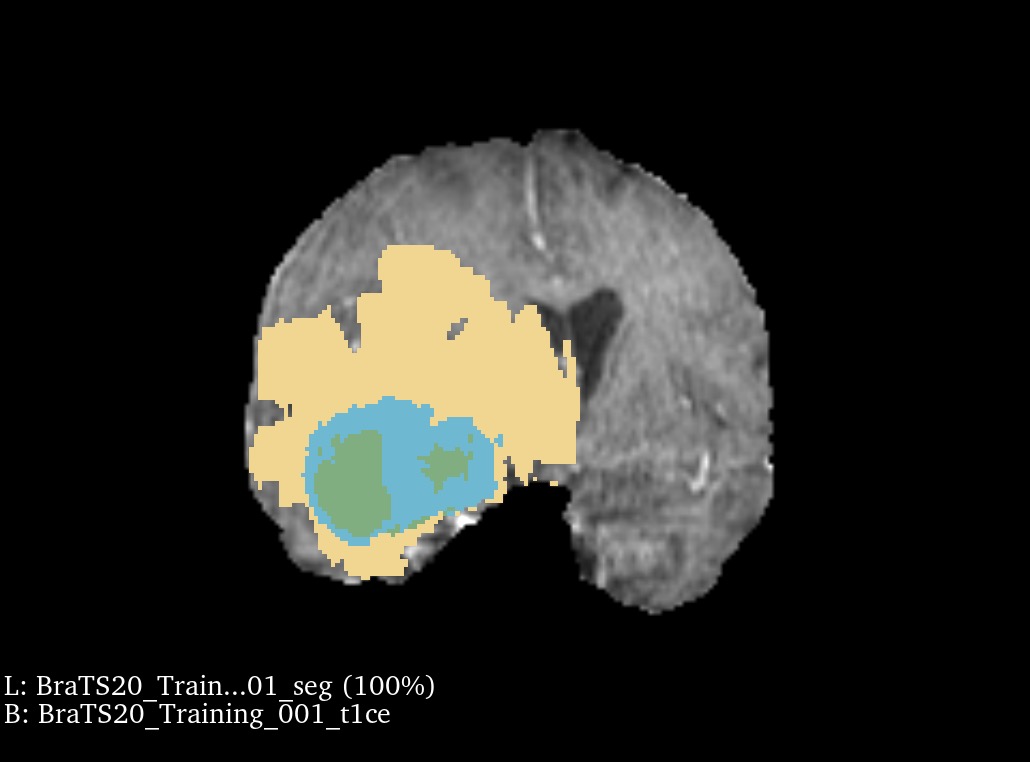}}
	\end{minipage}
	\caption{Visualization of one sample of BraTS 2020 dataset. We can observe significant variability in the target's shape, size, and distribution on each slice for a tumor target. Meanwhile, multiple-segmented targets are also present.}
	\label{fig:brats}
\end{figure}

\section{Introduction}
Brain tumors, caused by the abnormal growth of brain cells, pose a significant threat to human health, making early diagnosis and treatment crucial. MRI, as a non-invasive imaging technique, provides clear visualization of soft tissue lesions and is widely used in diagnosing and treating brain tumors, as shown in Figure \ref{fig:brats}. Current medical image segmentation methods primarily rely on U-shaped CNNs.

Despite extensive research, brain tumor segmentation remains challenging due to high variability in MRI images, unclear boundaries, and irregular tumor shapes and textures. Traditional CNN models struggle to adapt to these irregularities, failing to aggregate semantic information and compensate for spatial information loss. This leads to noise, misclassification, incomplete segmentation, limited image feature extraction, and constrained accuracy improvements.

Drawing from previous successful semantic segmentation studies, Guo et al. \cite{guo2022segnext} identified three key features, shown in Table \ref{tb:intro}, that a good CNN segmentation model should possess. We incorporated these key points into the brain tumor image segmentation characteristics and summarized them as follows:

\begin{table}[!t]
    \centering
    \caption{Three key features for semantic segmentation.}
    \label{tb:intro}
    \resizebox{0.9\columnwidth}{!}{%
        \begin{tabular}{c|ccc}
            \hline\hline
            \textbf{ } & \textbf{DLKA} & \textbf{SSPP} & \textbf{CAM} \\
            \hline
            \textbf{Strong Encoder} & $\checkmark$ & $\checkmark$ &  \\
            \textbf{Multi-Scale Interaction} &  & $\checkmark$ & $\checkmark$ \\
            \textbf{Attention Mechanism}s &  &  & $\checkmark$ \\
            \hline\hline
        \end{tabular}%
    }
\end{table}

(i) \textbf{Utilization of a powerful encoder.} Brain images typically encompass intricate structures such as brain tissue, vessels, and ventricles, while tumors often exhibit diverse shapes and sizes. A robust encoder is necessary to capture and represent these complex high-level semantic features, segmenting these structures accurately.

(ii) \textbf{Fusing multi-scale information.} Tumors within various organizational structures in the brain may exhibit significant size, shape, and distribution disparities. By fusing multi-scale information, the model can better capture details and global context in the image, enhancing the segmentation model's perception of various structures.

(iii) \textbf{Integration of attention mechanisms.} MRI images have multiple channels, each providing different information. Channel attention mechanisms help the model identify crucial channels for a specific task. Spatial attention mechanisms help the model focus on specific locations to capture local structural details, enhancing segmentation accuracy.


Inspired by Guo \cite{guo2022segnext}, we revisited CNN design principles to develop A4-Unet, a brain tumor segmentation architecture integrating four advanced components—Deformable Large Kernel Attention (DLKA), Swin-Enhanced Atrous Spatial Pyramid Pooling (SSPP), Combined Attention Module (CAM), and Attention Gates (AG) -- each enhancing performance. Our key innovations are:




\begin{itemize}
\item By incorporating large-kernel variable convolutions, the encoder can better capture multi-scale information with low complexity.

\item Long-distance dependencies intra-image and relationships inter-channel can be extracted by employing Swin Spatial Pyramid Pooling (SSPP) and convolutional channel attention in the bottleneck layer.

\item In the decoder, we leverage the orthogonality of Discrete Cosine Transform (DCT) to compute channel attention weights, followed by skip connections to supplement fine edge details. Additionally, we utilize simple convolutional element-wise multiplication to induce spatial attention, improving the generalization performance of a model.
\end{itemize}

\section{Related Work}
\subsection{Backbone Network}
\textbf{CNN-based Architecture.}
\sloppy
CNN-based methods classify pixel patches to capture local and global features. DenseNet \cite{Huang2017densenet} stacks deep layers to maintain multi-scale features, and Unet-based extensions \cite{Isen2021nnUNet}, inspired by Fully Convolutional Networks (FCNs), address various segmentation challenges. SegNeXt \cite{guo2022segnext} enhances convolutional structures with Multi-scale Convolutional Attention (MSCA) Module. However, despite effectively retaining low-level information, CNN models struggle to capture high-level information, limiting their performance.

\textbf{Transformer-based Networks.}
Transformer-based networks assign importance weights to image parts using attention mechanisms, Such networks have shown impressive results on vision tasks with the initial success of Vision Transformer (ViT) \cite{Dosovitskiy2020vit}. Variations like SegFormer \cite{xie2021segformer} and Swin Transformer \cite{liu2021swintrans} use hierarchical transformer encoders to extract multi-scale features with simple decoders for segmentation. However, they struggle with detecting high-resolution details like textures and edges, limiting their effectiveness in dense vision tasks.

\textbf{Integration of CNN and Transformer.}
Hybrid architectures combining CNNs and transformers leverage both strengths to overcome limitations. TransAttUnet \cite{chen2023transatt} integrates transformers and U-Net to capture global contextual information with attention blocks and multi-scale skip connections, achieving semantic consistency in feature maps. BoTNet \cite{Srinivas2021botnet} uses CNNs to process input images into tokenized feature maps, and then uses transformers to capture long-range dependencies. In our study, A4-Unet incorporates a robust convolutional encoder and transformer-guided modules to achieve a convincing segmentation performance.

\begin{figure*}[t]
    \centering
    \includegraphics[width=0.7\textwidth]{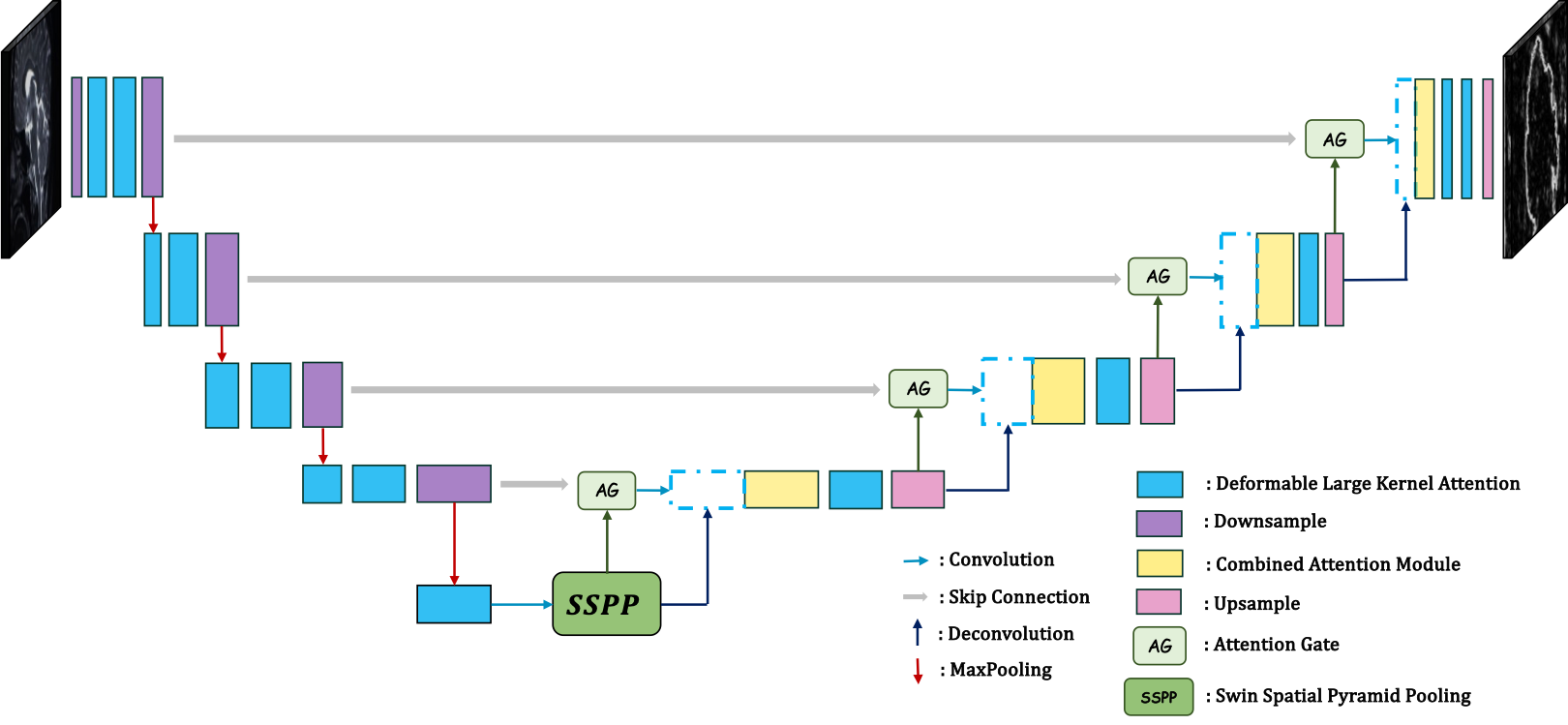}
    \caption{The overall architecture of our proposed A4-Unet.}
    \label{fig:a4unet}
\end{figure*}

\begin{figure}[!h]
        \centering
        \includegraphics[width=0.25\linewidth]{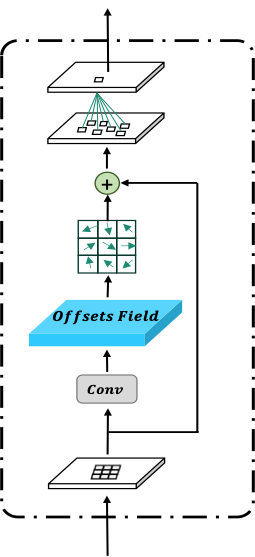}
        \caption{DLKA dynamically modifies convolutional weight coefficients and deformation offsets during training, enhancing the extraction of features from irregular objects in medical images.}
        \label{fig:dlka}
\end{figure}

\subsection{Attention Mechanisms}
Attention mechanisms dynamically adjust weights based on input features. Channel attention, like Squeeze-and-Excitation Network (SE-Net) \cite{hu2018squeeze}, assigns different weights to each channel, while Frequency Channel Attention Network (FcaNet) \cite{qin2021fcanet} uses Discrete Cosine Transformations to focus on low-frequency channel information.

Spatial attention enhances important regions by creating weight masks, as seen in Convolutional Block Attention Module (CBAM)\cite{woo2018cbam}, which combines pooling and concatenation for a unified feature descriptor. Our model integrates channel and spatial attention using CBAM's lightweight design to emphasize important regions and suppress irrelevant information, capturing cross-channel relationships and spatial details for precise detection.

\subsection{Adjustment of Receptive Field}
\textbf{Atrous Convolution.}
Atrous convolution first appeared in a dyadic wavelet transform technique \cite{holschneider1990real} that is well recognized as a signal processing technique. Deep networks reduce the final feature map resolution, resulting in the cumulative influence of pooling layers, striding operations, etc. Yu and Koltun \cite{yu2015multi} presented an innovative method to overcome this deficiency while seeking a more extensive information spectrum. 

\textbf{Deformable Convolution.}
CNNs' fixed receptive fields limit their ability to handle large-scale geometric transformations, making high-level semantic extraction challenging. Inspired by the multi-scale deformable part models\cite{Felz2009object} and spatial transformer module\cite{Jader2015spatial}, deformable convolution \cite{Dai2017deform} addresses this by introducing 2D offsets to sampling locations, allowing flexible grid deformation. We adopt deformable convolution to enhance receptive field flexibility for better target segmentation.

\subsection{Multi-scale Contextual Information}
\textbf{Atrous Spatial Pyramid Pooling.}
Aggregating multi-scale contextual information is crucial for accurate pixel-level classification in semantic segmentation. Dilated convolution \cite{yu2015dilate} enlarges the receptive field without changing output size. Building on SPP layers \cite{he2015spp}, ASPP \cite{chen2017deeplab} captures image context at multiple scales. This inspires our module to extract rich, comprehensive information from lesion images.

\textbf{Multi-scale Transformer.}
While CNNs have effectively used multi-scale feature representations, this potential has yet to be fully explored in vision transformers. CrossViT \cite{chen2021crossvit} introduces a dual-branch transformer with cross-attention, and MViT \cite{fan2021multiscale} embeds a multi-scale feature pyramid into the transformer. Inspired by these works, we propose a dual-branch encoder based on the hierarchical Swin transformer architecture.

\section{Methodology}
\subsection{Overall Architecture}
Our A4-Unet features an encoder-decoder architecture with three main components, as shown in Figure \ref{fig:a4unet}: DLKA for enhanced feature extraction, SSPP for multi-scale interactions, and CAM for attention mechanisms. The encoder uses DLKA, SSPP handles multi-scale features in the bottleneck, and the decoder aggregates features with gated and mixed attention across four upsampling stages, optimizing brain tumor segmentation.

\subsection{Strong Encoder}
To build a robust encoder, we integrate the Deformable Large Kernel Attention (DLKA) block in Figure~\ref{fig:dlka} into the downsampling process. DLKA includes a Deformable Convolution Module (DConv) and a Large Convolution Kernel (LK).

The DConv is ideal for enhancing low-level feature details like edges, textures, and shapes, particularly for medical targets with irregular sizes and various textures. The DConv consists of a 2D convolution, a Deformable Convolution with adjustable sampling grids using offsets, an activation function for nonlinearity, and an offset field calculation.  Proposed by Azad \cite{azad2023selfattention}, a standard convolution layer generates offsets, guiding the Deformable Convolution layer's sampling positions. The DConv module equation is as follows:

\begin{equation}
\label{eq:DConv1}
  Attention = Conv_{1 \times 1}(Conv_{DC}(Conv_{DW}(F)),
\end{equation}
\begin{equation}
\label{eq:DConv2}
  Output = Conv_{1 \times 1}(Attention \otimes F) + F,
\end{equation}

\noindent where $Conv_{DC}$ and $Conv_{DW}$ are deformable convolution and depth-wise
dilation convolution, respectively, while $F$ is the input feature.

On the other hand, although CNNs do well in capturing local features and low-level information, they come at the cost of neglecting the global context. The LK proposed by Guo et al. \cite{guo2023visual} can overcome this limitation by enlarging the receptive field. It provides a similar receptive field as the self-attention mechanism, with fewer parameters. The structure of LK contains a depth-wise convolution, a dilated convolution, and a $1 \times 1$ convolution. The kernel size of depth-wise convolution ($K_{DW}$) and dilated convolution ($K_{DC}$) can be calculated as below:
\begin{equation}
\label{eq:K_DW}
  K_{DW} = (2d-1) \times (2d-1),
\end{equation}
\begin{equation}
\label{eq:K_DC}
  K_{DC} = \left\lceil \frac{K}{d} \right\rceil \times \left\lceil \frac{K}{d} \right\rceil,
\end{equation}

\noindent where $d$ is dilation rate and $K$ is kernel size.

In sum, DLKA integrates into the encoder to provide long-range dependencies during downsampling and to concatenate with feature maps in the upsampling process via skip connections, thus compensating for low-level feature details.

\begin{figure}[t]
        \centering
        \includegraphics[width=\linewidth]{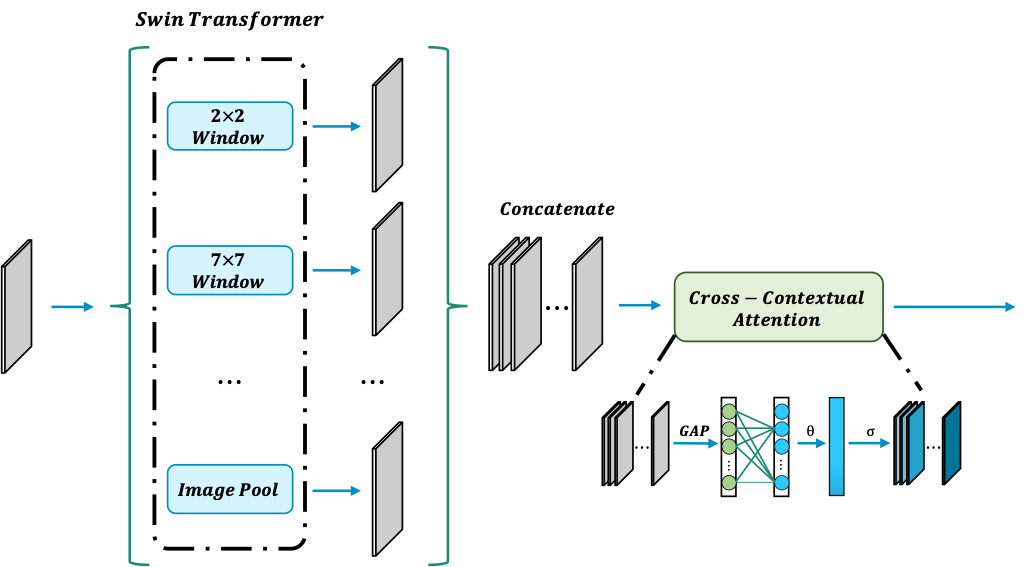}
        \caption{The implementation of SSPP and Cross-Contextual Attention module. The Swin Transformer uses small windows for local features and larger ones for global semantics. In the cross-attention block, multi-scale channel information is fused using an MLP layer and GAP to calculate attention scores.}
        \label{fig:sspp}
\end{figure}

\subsection{Multi-scale Interaction}
Addressing the challenges of irregular sizes and shapes in medical images requires introducing multi-scale interaction and enhancing spatial representation. Previous works \cite{ghafoorian2017location,hussain2018segmentation} used multi-scale patches and deeper networks, but multi-scale information remained fragmented.

We tackle this by modifying the bottleneck layer to include Swin Spatial Pyramid Pooling (SSPP) and a Cross-Contextual Attention module shown in Figure \ref{fig:sspp}. This approach integrates Swin Transformer blocks with varying window sizes, providing rich contextual information.

\begin{figure}[t]
        \centering
        \includegraphics[width=0.75\linewidth]{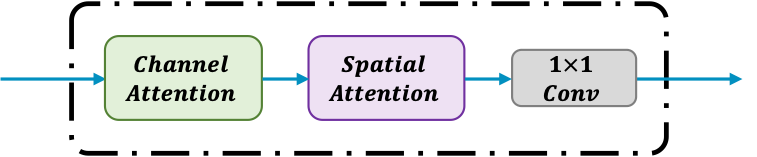}
        \caption{The general structure of Combined Attention Module. It consists of an orthogonal channel attention, a convolution-based spatial attention, and a $1 \times 1$ convolutional block.}
        \label{fig:cam}
\end{figure}

\textbf{Swin Spatial Pyramid Pooling.}
In DeepLab V3+, Chen et al. \cite{chen2017rethinking} introduced the Atrous Spatial Pyramid Pooling (ASPP) module, which dynamically selects convolutional blocks of varying sizes to handle different target scales. This approach prevents large targets from being fragmented and maintains long-distance dependencies without altering the network structure.

Inspired by SSPP by Azad et al. \cite{azad2022transdeeplab}, we replace four dilated convolutions with Swin Transformers to better capture long-range dependencies. The extracted features are merged and fed into a cross-contextual attention module. This enhances the model's ability to capture contextual dependencies across different scales.

\textbf{Cross-Contextual Attention.}
The ASPP concatenates feature maps via depth-wise separable convolution, which does not capture channel dependencies. To address this, Azad introduced cross-contextual attention after SSPP feature fusion. Assume each SSPP layer has tokens ($P$) and embedding dimension ($C$) as ($z_{m}^{P \times C}$), representing objects at different scales. We create a multi-scale representation $z_{all}^{P \times MC} = [z_1||z_2...||z_M]$ by concatenating these features. A scale attention module then emphasizes each feature map's contribution, using global representation and an MLP layer to generate scaling coefficients ($w_{scale}$), enhancing contextual dependencies:

\begin{equation}
\label{eq
}
w_{scale} = \sigma(W_2\delta(W_1GAP_{z_{all}})),
\end{equation}
\begin{equation}
\label{eq
}
z_{all}' = w_{scale} \cdot z_{all},
\end{equation}

\noindent where $W_1$ and $W_2$ are learnable MLP parameters, $\delta$ is the ReLU function, $\sigma$ is the Sigmoid function, and GAP is global average pooling.

In the second attention level, Cross-Contextual Attention learns scaling parameters to enhance informative tokens by calculating their weight maps, using the same strategy:

\begin{equation}
\label{eq
}
w_{tokens} = \sigma(W_3\delta(W_4GAP_{z_{all}'})),
\end{equation}
\begin{equation}
\label{eq
}
z_{all}'' = w_{tokens} \cdot z_{all}',
\end{equation}


\subsection{Convolutional Attention Module}
We construct our decoder by integrating a novel convolutional attention module with a frequency feature that effectively suppresses unnecessary information. Furthermore, we introduce skip connections with attention-gated fusion, contributing to the suppression of irrelevant regions and accentuation of salient features.

As shown in Figure \ref{fig:cam}, our decoder includes a vanilla block for feature upsampling, an Attention Gate (AG) for cascaded feature fusion, and a Combined Attention Module (CAM) for feature map enhancement. We use four CAM blocks for the four pyramid layers of the encoder and four AGs for skip connections. Multi-scale features are consolidated by combining upsampled features from the previous layer with skip connection features using AG. The CAM module then enhances pixel grouping and suppresses background information with frequency channel and spatial attention (SA). Finally, Dconv propagates the fused features to the upper layer.

\subsubsection{\textbf{Combined Attention Module}}
\begin{figure}[!t]
        \centering
        \includegraphics[width=\linewidth]{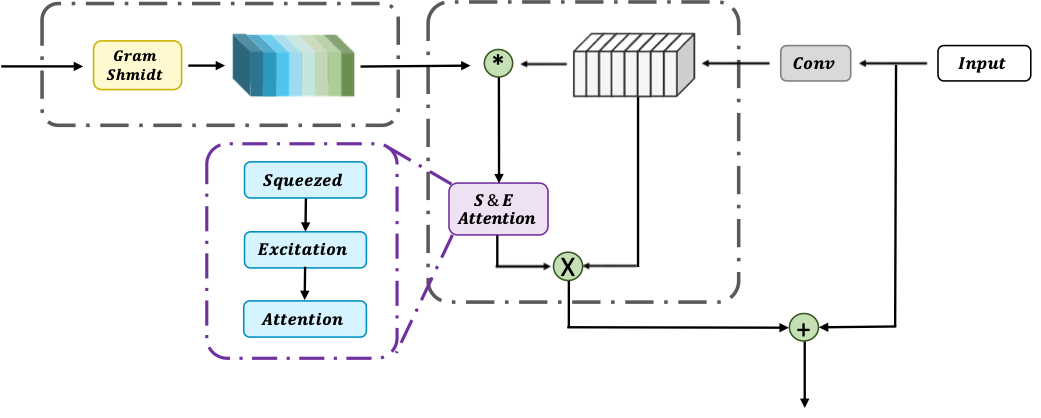}
        \caption{The implementation of Channel Attention in the CAM.}
        \label{fig:ca}
\end{figure}

\begin{itemize}
    \item \textbf{Channel Attention}
\end{itemize}
To enhance channel attention accuracy in CAM, we replaced convolution-based channel attention with Orthogonal Channel Attention (OCA) from Salman et al. \cite{salman2023orthonets}. OrthoNet's channel attention addresses the limitation of Global Average Pooling (GAP) by using the Discrete Cosine Transform (DCT) to preserve low-frequency information. As shown in Figure \ref{fig:ca}, OCA's structure involves selecting suitable filters within appropriate dimensions and ensuring filter orthogonality using the Gram-Schmidt process. This structure enhances feature representation in neural networks.

\begin{figure}[t]
        \centering
        \includegraphics[width=0.9\linewidth]{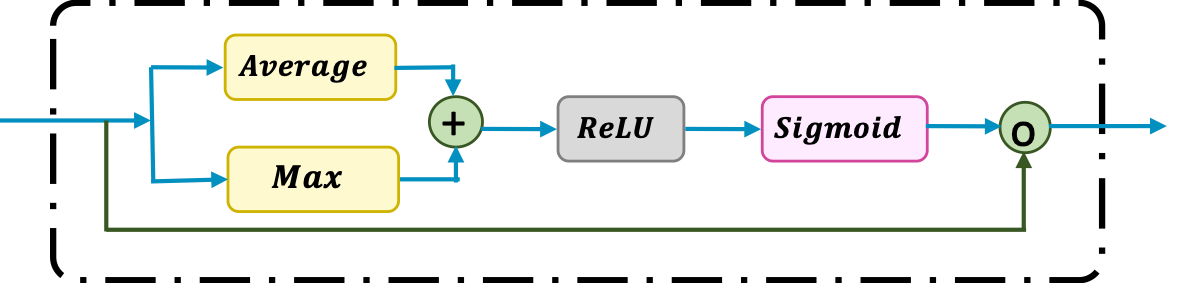}
        \caption{The implementation of Spatial Attention in the CAM. It uses convolution to represent the maximum and average values obtained along the channel dimension. And $1 \times 1$ convolution for fusing information across various channels.}
        \label{fig:sa}
\end{figure}

\begin{itemize}
    \item \textbf{Spatial Attention}
\end{itemize}
Spatial attention helps the model adapt to spatial variability by adjusting attention to local structures, improving generalization. As shown in Figure \ref{fig:sa}, for each feature point in input feature $F$ of size $H*W$, the maximum and average values along the channel axis are denoted as $F_{max} \in R^{1*H*W}$ and $F_{avg} \in R^{1*H*W}$, and concatenated into a $2*H*W$ tensor. This tensor undergoes convolution to create a spatial attention map that highlights or suppresses specific locations.
\begin{equation}
    SA = Conv(MaxPool(F), AvgPool(F))
\end{equation}

\subsubsection{\textbf{Attention Gate}}

\begin{figure}
        \centering
        \includegraphics[width=\linewidth]{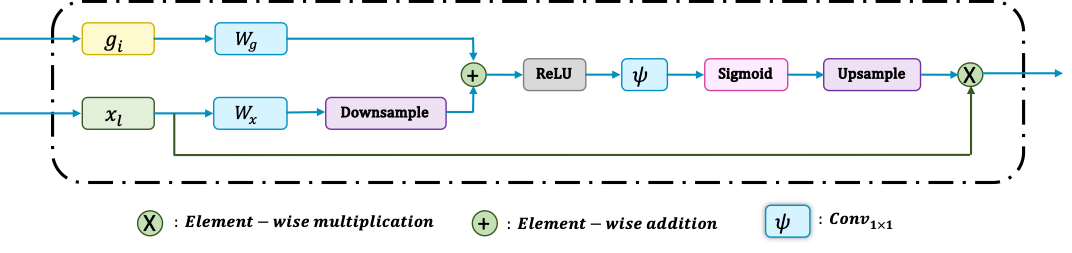}
        \caption{Attention Gates (AGs) use gating from coarser scales to filter features transmitted through skip connections, achieving feature selectivity. The input image is incrementally filtered and downsampled by a factor of 2 at each encoder scale (e.g., $H_4=(H_1)/8$) where $N_c$ represents the number of classes.}
        \label{fig:ag}
\end{figure}

We incorporate the attention gate into the skip connection process. Figure \ref{fig:ag} illustrates the architecture of an attention gate unit. Let $x_l$ represent the feature map of layer $l$. For each pixel $i$, a gating signal $g_i$ vector is used to identify focal areas at a larger scale. The coefficient of attention, denoted as $\alpha$, ranges from 0 to 1, selecting relevant feature responses and suppressing irrelevant feature details. The resulting $x_{output}$ is obtained through element-wise multiplication of $x_l$ and $\alpha$, calculated as follows:
\begin{equation}
x_{output} = x_l \cdot \alpha_i
\end{equation}

According to the formula, the gating coefficient $\alpha$ is derived through additive attention. Given the complexity of medical images involving multiple semantic classes, we incorporate the multi-dimensional attention coefficient \cite{shen2018disan} to concentrate on target regions. The computation of the multi-dimensional attention coefficient involves the following:
\begin{equation}
  \alpha_i = \sigma(\Psi^T(\delta(W_x^Tx_l+W_g^Tg_i+b_g)) + b_\Psi)
\end{equation}

\noindent where $W_x$, $W_g$ are bias, $\sigma(x) = \frac{a}{1+e^{-x}}$ is the Sigmoid function and $\sigma(x) = max(0, x)$ is the ReLU function. As for gating signal vector $g_i$, we adopt $1 \times 1$ channel-wise convolution (represented as $\Psi$ in the formula) as the linear transformation on the feature map $x_l$.

\section{Results}
In this section, we first conduct comprehensive ablation studies to validate the effectiveness of our design. Then, we compare our results with several state-of-the-art networks and analyze the reasons for the results.

\subsection{Dataset}
The BraTS datasets are part of the Brain Tumor Segmentation Challenge. We select the BraTS 2019, 2020, and 2021 datasets as the experimental data for our study. These are publicly available via the following links\footnote{\href{https://www.med.upenn.edu/cbica/brats-2019/data.html}{BraTS 2019}, \href{https://www.med.upenn.edu/cbica/brats2020/data.html}{BraTS 2020}, \href{https://www.synapse.org/\#!Synapse:syn25829067/wiki/610863}{BraTS 2021}}. All BraTS multimodal scans are provided as NIfTI files (.nii.gz) and include the following: I) native T1-weighted scans (T1N), II) post-contrast T1-weighted scans (T1C/T1CE, also referred to as T1Gd), III) T2-weighted scans (T2W/T2), and IV) T2 Fluid Attenuated Inversion Recovery scans (T2F/FLAIR). The training and validation sets have unspecified glioma classifications, and all data underwent standardized preprocessing by the challenge organizers.

In addition to public benchmarks, we evaluated our model on a proprietary dataset from an anonymous institution. This dataset includes T1c and T2 MRI images from 194 glioma patients, annotated for whole tumors by senior radiologists. Since our model is 2D, we sliced each 3D MRI image into 2D slices. Details are shown in Table \ref{tb:data}.

\begin{table}[!h]
    \centering
    \caption{Details of datasets.}
    \label{tb:data}
    \resizebox{0.5\textwidth}{!}{%
        \begin{tabular}{c|cccc}
            \hline\hline
            \textbf{ } & \textbf{BraTS 2019} & \textbf{BraTS 2020} & \textbf{BraTS 2021} & \textbf{Proprietary Dataset} \\
            \hline
            \textbf{Training set} & 335 & 369 & 1251 & 155 \\
            \textbf{Validation set} & 125 & 125 & 219 & 39 \\
            \textbf{Testing set} & 166 & 166 & 570 & - \\
            \textbf{Modalities} & flair, t1ce, t1, t2  & flair, t1ce, t1, t2 & t1n, t1c, t2w, t2f & t1c, t2 \\
            \textbf{Slices} & 51,925 & 57,195 & 193,905 & 7,760 \\
            \hline\hline
        \end{tabular}%
    }
\end{table}

\subsection{Metrics}
\label{sec:mtc}
\subsubsection{\textbf{Dice Similarity Coefficient}}
\
\newline
The Dice Similarity Coefficient (DSC) is a key metric for evaluating segmentation models, ranging from 0 to 1 to represent similarity between two samples. It is calculated as:
\begin{equation}
DSC = \frac{2TP}{FN+FP+2TP}
\end{equation}
Here, $TP$ represents true positive pixels, $FP$ indicates false positive pixels, and $FN$ represents false negative pixels.

\subsubsection{\textbf{Mean Intersection over Union}}
\
\newline
The IoU calculates the intersection of the predicted and true segmentation divided by their union. As an extension, the mIoU computes the IoU for each class and then calculates the mean of these IoU scores. The mIoU provides a more comprehensive assessment of the overall segmentation performance across $k$ different classes.
\begin{equation}
  mIoU = \frac{1}{k+1}\sum_{i=0}^{k}\frac{TP}{FN+FP+TP}
\end{equation}

\subsubsection{\textbf{Hausdorff Distance}}
\
\newline
Hausdorff Distance (HD) measures the maximum distance from each point in the predicted boundary set to its nearest point in the ground truth boundary set, assessing segmentation accuracy by comparing boundary correspondence. Given sets $A$ (predicted) and $B$ (ground truth), the Hausdorff distance formula is:
\begin{equation}
  H(A,B) = max(h(A,B),h(B,A))
\end{equation}
where 
\begin{equation}
  h(A,B) = \max_{a \in A} \min_{b \in B} \|a - b\|
\end{equation}
\begin{equation}
  h(B,A) = \max_{b \in B} \min_{a \in A} \|b - a\|
\end{equation}

\subsection{Implementation Details}
All experiments are implemented in PyTorch 2.0.1 and trained on a single GeForce GTX 4090 GPU with 24 GB memory. We use standard back-propagation with the AdamW optimizer and Softmax activation function. Training employs a batch size of 16, an initial learning rate of 1e-5, and runs for 30 epochs. Total training time varies by dataset size: approximately 20 hours for BraTS 2019, 30 hours for BraTS 2020, and 50 hours for BraTS 2021.

\subsection{Ablation Study}
We conducted an ablation study on the BraTS 2020 dataset to analyze the effectiveness of three crucial factors. Results are shown in Table \ref{tb:ab2}. We observed that the BraTS 2019 dataset had slower convergence, requiring 12 epochs compared to 10 epochs for the other two datasets, likely due to its smaller training sample size.

\subsubsection{\textbf{Effect of the Strong Encoder}}
To validate the effect of DLKA in the encoder, we construct the baseline network and another version with DLKA. Employing the DLKA module leads to an improvement in the Dice score of 1.3\% compared to the baseline. It also demonstrates a slight improvement when combined with other blocks (e.g., SSPP, CAM).

\subsubsection{\textbf{Effect of the Multi-Scale Interaction}}
We evaluated the SSPP block for multi-scale information fusion and found a 2.0\% accuracy improvement over the baseline. Compared to DLKA, the SSPP module had a more significant impact on accuracy, demonstrating that the transformer can better capture global features. This highlights the importance of introducing a global context for brain tumor segmentation.

\subsubsection{\textbf{Effect of the CAM}}
As for the CAM block in the decoder, we can conclude that the attention mechanisms result in a 1.9\% improvement in model performance, as shown in Table \ref{tb:ab2}. When the CAM fusion with DLKA, the model can achieve a better result, adequately demonstrating the effectiveness of adopting skip connections using DLKA before the CAM block.

\begin{table}[t]
    \centering
    \caption{Ablation Study on Dice Score (\%).}
    \label{tb:ab2}
    \resizebox{0.5\textwidth}{!}{%
        \begin{tabular}{c|ccc}
            \hline\hline
            \textbf{Model} & \textbf{BraTS19} & \textbf{BraTS20} & \textbf{BraTS21} \\
            \hline
            ResUnet \textbf{(baseline)} & 92.58 & 92.22 & 91.67 \\
            ResUnet + DLKA & 93.07 & 93.48 & 91.70 \\
            ResUnet + SSPP & 93.43 & 94.12 & 92.01 \\
            ResUnet + CAM & 93.92 & 94.05 & 91.91 \\
            ResUnet + DLKA + SSPP & 93.51 & 94.38 & 92.30 \\
            ResUnet + DLKA + CAM & 94.07 & 93.74 & 92.34 \\
            ResUnet + SSPP + CAM & 94.19 & 94.12 & 92.56 \\
            \textbf{A4-Unet (ours)} & \textbf{94.61} & \textbf{94.47} & \textbf{92.84} \\
            \hline\hline
        \end{tabular}%
    }
\end{table}

\begin{table}[t]
    \centering
    \caption{Comparisons on different metrics using A4-Unet.}
    \label{tb:ab1}
    \resizebox{0.5\textwidth}{!}{%
        \begin{tabular}{c|ccc}
            \hline\hline
            \textbf{ } & \textbf{DSC (\%) $\uparrow$} & \textbf{mIoU (\%) $\uparrow$} & \textbf{HD95 (mm) $\downarrow$} \\
            \hline
            \textbf{Proprietary Dataset} & 84.18 & 81.60 & 10.77 \\
            \textbf{BraTS 2019} & 94.61 & 99.85 & 13.34 \\
            \textbf{BraTS 2020} & 94.47 & 99.68 & 8.57 \\
            \textbf{BraTS 2021} & 92.84 & 99.89 & 12.50 \\
            \hline\hline
        \end{tabular}%
    }
\end{table}

\subsection{Quantitative Analysis and Visualization}

\begin{figure}[!h]
        \centering
        \includegraphics[width=1\linewidth]{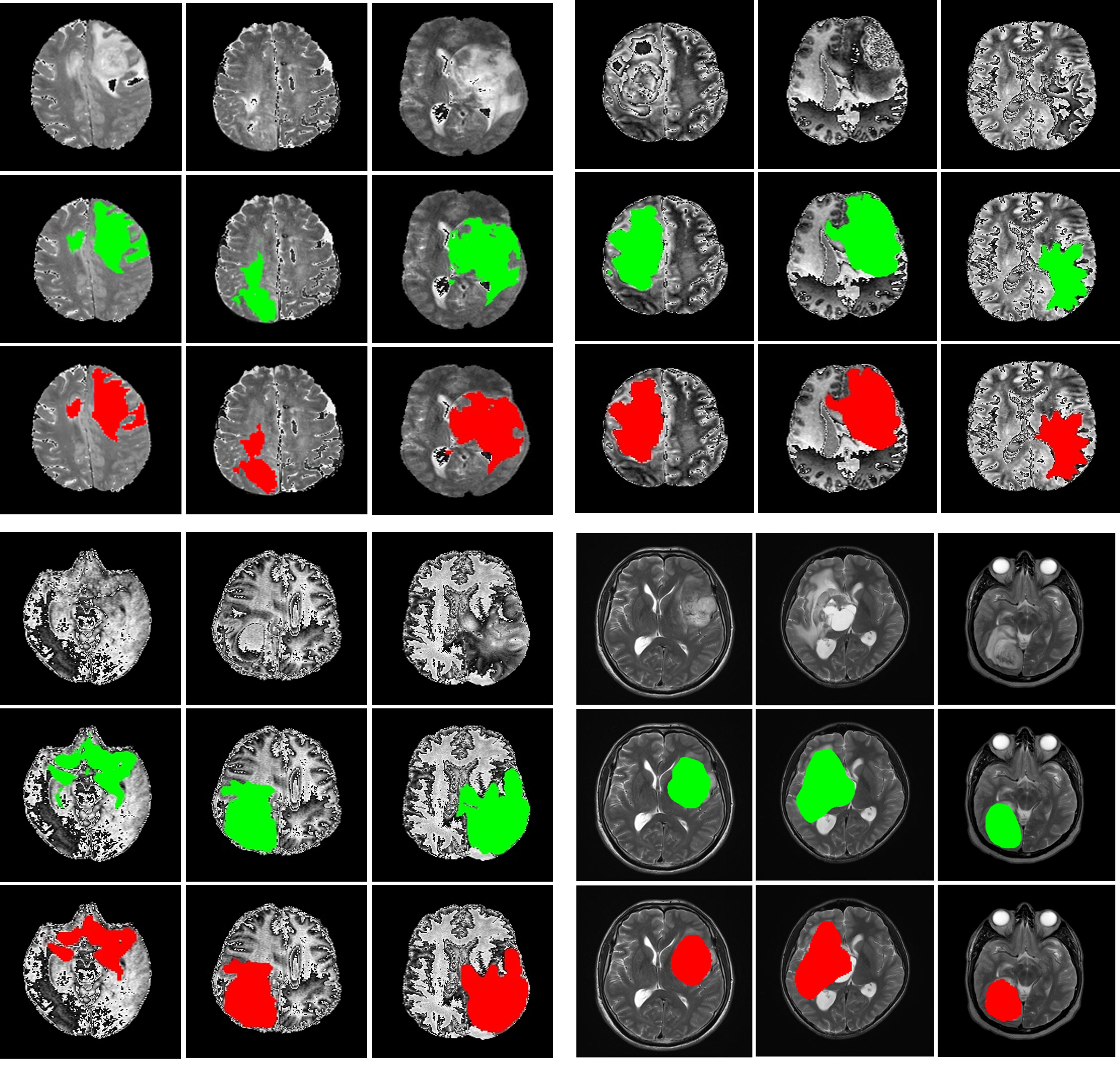}
        \caption{From top left to bottom right are the visualizations of the BraTS19 dataset, BraTS20 dataset, BraTS21 dataset, and our proprietary dataset, respectively. Green mask represents the ground truth, while red mask represents our results.}
        \label{fig:visa}
\end{figure}

We test the proposed A4-Unet by evaluating three metrics mentioned in Section~\ref{sec:mtc} on BraTS 2019, BraTS 2020, and BraTS 2021 datasets, respectively. The experimental results on each training dataset represent the average of five independent runs and were subjected to cross-validation. The results are described in Table \ref{tb:ab1}, and the visualization is illustrated in Figure \ref{fig:visa}. We got a lower HD95 score of 8.57 on the BraTS 2020 than the other two datasets. We attribute this improvement primarily to two reasons: (i) The BraTS 2020 dataset contains larger segmentation targets, and our model has higher segment performance than irregular and small targets. (ii) 95\% might not be the optimal hyperparameter for the more complex datasets like the BraTS 2019 and BraTS 2021, leading to differences in their results.

On a proprietary dataset, our model achieved a Dice coefficient of 84.18\%, mIoU of 81.60\%, and HD95 of 10.77mm, lower than on BraTS datasets. This was attributed to the proprietary dataset having fewer modalities and tumor features, limiting the model's ability to learn optimal features.

\subsection{Comparisons}
The proposed A4-Unet model follows the standard CNN segmentation network design paradigm. To evaluate its improvements and component effectiveness, we compared it with state-of-the-art networks on the three BraTS datasets. Comparative results are cited from the literature. Since official ranking criteria consider multiple metrics, a challenge champion's DSC score may not be the highest. Results are shown in Table \ref{tab:compare}.

\noindent\textbf{BraTS 2019.} We compared A4-Unet with four models on the BraTS 2019 dataset, showing a 21.03\% and 20.53\% Dice Score improvement over TransUnet and Swin-Unet, respectively. Unlike the transformer-based models requiring more parameters and data, A4-Unet uses DLKA for an efficient encoder. It also surpasses Cascade Unet by integrating attention mechanisms and multiscale fusion, enhancing fine-edge detail through Attention Gates, thus improving segmentation performance.

\noindent\textbf{BraTS 2020.} On the BraTS 2020 dataset Table \ref{tab:compare}, A4-Unet achieved a Dice score of 94.47\%, mIoU of 99.68\%, and 95th percentile Hausdorff distance of 8.57mm, outperforming Swin-Unet, TransUnet, nnUnet \cite{Isen2021nnUNet}, and ResUnet+. TransUnet and Swin-Unet faced similar issues due to dataset size. nnUnet won the BraTS 2020 challenge with only targeted training and post-processing. Compared to the strategy of nnUnet, we focused on improving the network architecture and achieved significant enhancements.

\begin{table}[!ht]
    \centering
    \caption{Performance Comparisons on BraTS Datasets.}
    \label{tab:compare}
    \renewcommand{\arraystretch}{1.2}
    \resizebox{0.5\textwidth}{!}{%
        \begin{tabular}{c|l|c|c|c}
            \hline\hline
            \textbf{Challenge} & \textbf{Method} & \textbf{DSC (\%) $\uparrow$} & \textbf{mIoU (\%) $\uparrow$} & \textbf{HD95 (mm) $\downarrow$} \\
            \hline
            \textbf{BraTS} & TransUnet\cite{chen2021transunet} & 78.17 & - & 6.92 \\
            \textbf{2019} & Swin-Unet\cite{cao2022swin} & 78.49 & - & 4.83 \\
            \textbf{ } & ResUnet+\cite{resunet+} & 88.30 & 92.38 & - \\
            \textbf{ } & Cascade Unet\cite{tscunet} & 88.80 & - & \textbf{4.62} \\
            \textbf{ } & \textbf{A4-Unet (Ours)} & \textbf{94.61} & \textbf{99.85} & 13.34 \\
            \hline
            \textbf{BraTS} & Swin-Unet\cite{cao2022swin} & 89.34 & - & 11.1 \\
            \textbf{2020} & TransUnet\cite{chen2021transunet} & 89.46 & - & 12.85 \\
            \textbf{ } & nnUnet\cite{Isen2021nnUNet} & 91.18 & - & \textbf{8.49} \\
            \textbf{ } & ResUnet+\cite{resunet+} & 92.80 & 92.42 & - \\
            \textbf{ } & \textbf{A4-Unet (Ours)} & \textbf{94.47} & \textbf{99.68} & 8.57 \\
            \hline
            \textbf{BraTS} & UNETR\cite{hatamizadeh2022unetr} & 91.11 & - & - \\
            \textbf{2021} & Swin UNETR\cite{hatamizadeh2022swin} & 92.61 & - & 5.30 \\
            \textbf{ } & SegResNet\cite{siddiquee2021redundancy} & 92.65 & - & \textbf{3.60} \\
            \textbf{ } & Optimized Unet\cite{futrega2021optimized} & 92.68 & - & - \\
            \textbf{ } & Coupling nnUnet\cite{couplingunet} & 92.83 & - & 3.76 \\ 
            \textbf{ } & \textbf{A4-Unet (Ours)} & \textbf{92.84} & \textbf{99.89} & 12.50 \\
            \hline\hline
        \end{tabular}%
    }
\end{table}

\noindent\textbf{BraTS 2021.} For the BraTS 2021 dataset, we compared A4-Unet with UNETR\cite{hatamizadeh2022unetr}, Swin UNETR\cite{hatamizadeh2022swin}, SegResNet\cite{siddiquee2021redundancy}, Optimized Unet\cite{futrega2021optimized}, and Coupling nnUnet\cite{couplingunet}. While UNETR and Swin UNETR's transformer-based encoders increase parameters and training difficulty, A4-Unet's DLKA maintains low complexity and superior performance with stable parameters. SegResNet's dense skip connections are enhanced in our network by using attention gates to better utilize edge detail information for fine segmentation masks.

\subsection{Discussion}
Despite excellent performance on public datasets, the model still faces challenges in clinical applications. The diversity and complexity of real-world clinical data, (e.g., our proprietary dataset) complicate feature extraction and model learning, while the limited annotated data constrain the model's generalization capabilities. Therefore, further improvements are necessary before the model can be effectively applied in clinical settings.

\section{Conclusions}
In this paper, we presented A4-Unet, a brain tumor segmentation network that introduces Deformable Kernel Large Convolution (DLKA), Swin Spatial Pyramid Pooling (SSPP), and attention mechanisms, all while maintaining relatively low network complexity. This approach enables efficient multi-scale feature extraction, captures long-range dependencies, and integrates high-level and low-level semantic information. Our comparative experiments across three datasets demonstrate that A4-Unet significantly outperforms several state-of-the-art models, setting new benchmarks in segmentation performance. Notably, our model achieved substantial improvements in Dice Score and mIoU.


\section*{Acknowledgements}
This work is supported by Guangdong Provincial Key Laboratory of IRADS, BNU-HKBU United International College (2022B1212010006, R0400001-22), Guangdong Basic and Applied Basic Research Foundation (2024A1515011274), Guangdong Province General Universities Key Field Project (New Generation Information Technology) (2023ZDZX1033), and UIC Research Grant (UICR04202401-21).

\bibliographystyle{IEEEtran}
\bibliography{reference}

\begin{thebibliography}{10}
\providecommand{\url}[1]{#1}
\csname url@samestyle\endcsname
\providecommand{\newblock}{\relax}
\providecommand{\bibinfo}[2]{#2}
\providecommand{\BIBentrySTDinterwordspacing}{\spaceskip=0pt\relax}
\providecommand{\BIBentryALTinterwordstretchfactor}{4}
\providecommand{\BIBentryALTinterwordspacing}{\spaceskip=\fontdimen2\font plus
\BIBentryALTinterwordstretchfactor\fontdimen3\font minus \fontdimen4\font\relax}
\providecommand{\BIBforeignlanguage}[2]{{%
\expandafter\ifx\csname l@#1\endcsname\relax
\typeout{** WARNING: IEEEtran.bst: No hyphenation pattern has been}%
\typeout{** loaded for the language `#1'. Using the pattern for}%
\typeout{** the default language instead.}%
\else
\language=\csname l@#1\endcsname
\fi
#2}}
\providecommand{\BIBdecl}{\relax}
\BIBdecl

\bibitem{guo2022segnext}
M.-H. Guo, C.-Z. Lu, Q.~Hou, Z.~Liu, M.-M. Cheng, and S.-M. Hu, ``Segnext: Rethinking convolutional attention design for semantic segmentation,'' \emph{Advances in Neural Information Processing Systems}, vol.~35, pp. 1140--1156, 2022.

\bibitem{Huang2017densenet}
G.~Huang, Z.~Liu, L.~Van Der~Maaten, and K.~Q. Weinberger, ``Densely connected convolutional networks,'' in \emph{Proceedings of the IEEE conference on computer vision and pattern recognition}, 2017, pp. 4700--4708.

\bibitem{Isen2021nnUNet}
F.~Isensee, P.~Jaeger, S.~Kohl, J.~Petersen, and K.~Maier-Hein, ``nnu-net: a self-configuring method for deep learning-based biomedical image segmentation,'' \emph{Nature Methods}, vol.~18, pp. 1--9, 02 2021.

\bibitem{Dosovitskiy2020vit}
A.~Dosovitskiy, L.~Beyer, A.~Kolesnikov, D.~Weissenborn, X.~Zhai, T.~Unterthiner, M.~Dehghani, M.~Minderer, G.~Heigold, S.~Gelly \emph{et~al.}, ``An image is worth 16x16 words: Transformers for image recognition at scale,'' \emph{arXiv preprint arXiv:2010.11929}, 2020.

\bibitem{xie2021segformer}
E.~Xie, W.~Wang, Z.~Yu, A.~Anandkumar, J.~M. Alvarez, and P.~Luo, ``Segformer: Simple and efficient design for semantic segmentation with transformers,'' \emph{Advances in neural information processing systems}, vol.~34, pp. 12\,077--12\,090, 2021.

\bibitem{liu2021swintrans}
Z.~Liu, Y.~Lin, Y.~Cao, H.~Hu, Y.~Wei, Z.~Zhang, S.~Lin, and B.~Guo, ``Swin transformer: Hierarchical vision transformer using shifted windows,'' in \emph{Proceedings of the IEEE/CVF international conference on computer vision}, 2021, pp. 10\,012--10\,022.

\bibitem{chen2023transatt}
B.~Chen, Y.~Liu, Z.~Zhang, G.~Lu, and A.~W.~K. Kong, ``Transattunet: Multi-level attention-guided u-net with transformer for medical image segmentation,'' \emph{IEEE Transactions on Emerging Topics in Computational Intelligence}, 2023.

\bibitem{Srinivas2021botnet}
A.~Srinivas, T.-Y. Lin, N.~Parmar, J.~Shlens, P.~Abbeel, and A.~Vaswani, ``Bottleneck transformers for visual recognition,'' in \emph{Proceedings of the IEEE/CVF conference on computer vision and pattern recognition}, 2021, pp. 16\,519--16\,529.

\bibitem{hu2018squeeze}
J.~Hu, L.~Shen, and G.~Sun, ``Squeeze-and-excitation networks,'' in \emph{Proceedings of the IEEE conference on computer vision and pattern recognition}, 2018, pp. 7132--7141.

\bibitem{qin2021fcanet}
Z.~Qin, P.~Zhang, F.~Wu, and X.~Li, ``Fcanet: Frequency channel attention networks,'' in \emph{Proceedings of the IEEE/CVF international conference on computer vision}, 2021, pp. 783--792.

\bibitem{woo2018cbam}
S.~Woo, J.~Park, J.-Y. Lee, and I.~S. Kweon, ``Cbam: Convolutional block attention module,'' in \emph{Proceedings of the European conference on computer vision (ECCV)}, 2018, pp. 3--19.

\bibitem{holschneider1990real}
M.~Holschneider, R.~Kronland-Martinet, J.~Morlet, and P.~Tchamitchian, ``A real-time algorithm for signal analysis with the help of the wavelet transform,'' in \emph{Wavelets: Time-Frequency Methods and Phase Space Proceedings of the International Conference, Marseille, France, December 14--18, 1987}.\hskip 1em plus 0.5em minus 0.4em\relax Springer, 1990, pp. 286--297.

\bibitem{yu2015multi}
F.~Yu and V.~Koltun, ``Multi-scale context aggregation by dilated convolutions,'' \emph{arXiv preprint arXiv:1511.07122}, 2015.

\bibitem{Felz2009object}
P.~F. Felzenszwalb, R.~B. Girshick, D.~McAllester, and D.~Ramanan, ``Object detection with discriminatively trained part-based models,'' \emph{IEEE Transactions on Pattern Analysis and Machine Intelligence}, vol.~32, no.~9, pp. 1627--1645, 2010.

\bibitem{Jader2015spatial}
M.~Jaderberg, K.~Simonyan, A.~Zisserman \emph{et~al.}, ``Spatial transformer networks,'' \emph{Advances in neural information processing systems}, vol.~28, 2015.

\bibitem{Dai2017deform}
J.~Dai, H.~Qi, Y.~Xiong, Y.~Li, G.~Zhang, H.~Hu, and Y.~Wei, ``Deformable convolutional networks,'' in \emph{Proceedings of the IEEE international conference on computer vision}, 2017, pp. 764--773.

\bibitem{yu2015dilate}
F.~Yu and V.~Koltun, ``Multi-scale context aggregation by dilated convolutions,'' \emph{arXiv preprint arXiv:1511.07122}, 2015.

\bibitem{he2015spp}
K.~He, X.~Zhang, S.~Ren, and J.~Sun, ``Spatial pyramid pooling in deep convolutional networks for visual recognition,'' \emph{IEEE transactions on pattern analysis and machine intelligence}, vol.~37, no.~9, pp. 1904--1916, 2015.

\bibitem{chen2017deeplab}
L.-C. Chen, G.~Papandreou, I.~Kokkinos, K.~Murphy, and A.~L. Yuille, ``Deeplab: Semantic image segmentation with deep convolutional nets, atrous convolution, and fully connected crfs,'' \emph{IEEE transactions on pattern analysis and machine intelligence}, vol.~40, no.~4, pp. 834--848, 2017.

\bibitem{chen2021crossvit}
C.-F.~R. Chen, Q.~Fan, and R.~Panda, ``Crossvit: Cross-attention multi-scale vision transformer for image classification,'' in \emph{Proceedings of the IEEE/CVF international conference on computer vision}, 2021, pp. 357--366.

\bibitem{fan2021multiscale}
H.~Fan, B.~Xiong, K.~Mangalam, Y.~Li, Z.~Yan, J.~Malik, and C.~Feichtenhofer, ``Multiscale vision transformers,'' in \emph{Proceedings of the IEEE/CVF international conference on computer vision}, 2021, pp. 6824--6835.

\bibitem{azad2023selfattention}
R.~Azad, L.~Niggemeier, M.~H{\"u}ttemann, A.~Kazerouni, E.~K. Aghdam, Y.~Velichko, U.~Bagci, and D.~Merhof, ``Beyond self-attention: Deformable large kernel attention for medical image segmentation,'' in \emph{Proceedings of the IEEE/CVF Winter Conference on Applications of Computer Vision}, 2024, pp. 1287--1297.

\bibitem{guo2023visual}
M.-H. Guo, C.-Z. Lu, Z.-N. Liu, M.-M. Cheng, and S.-M. Hu, ``Visual attention network,'' \emph{Computational Visual Media}, vol.~9, no.~4, pp. 733--752, 2023.

\bibitem{ghafoorian2017location}
M.~Ghafoorian, N.~Karssemeijer, T.~Heskes, I.~W. van Uden, C.~I. Sanchez, G.~Litjens, F.-E. de~Leeuw, B.~van Ginneken, E.~Marchiori, and B.~Platel, ``Location sensitive deep convolutional neural networks for segmentation of white matter hyperintensities,'' \emph{Scientific Reports}, vol.~7, no.~1, p. 5110, 2017.

\bibitem{hussain2018segmentation}
S.~Hussain, S.~M. Anwar, and M.~Majid, ``Segmentation of glioma tumors in brain using deep convolutional neural network,'' \emph{Neurocomputing}, vol. 282, pp. 248--261, 2018.

\bibitem{chen2017rethinking}
L.-C. Chen, G.~Papandreou, F.~Schroff, and H.~Adam, ``Rethinking atrous convolution for semantic image segmentation,'' \emph{arXiv preprint arXiv:1706.05587}, 2017.

\bibitem{azad2022transdeeplab}
R.~Azad, M.~Heidari, M.~Shariatnia, E.~K. Aghdam, S.~Karimijafarbigloo, E.~Adeli, and D.~Merhof, ``Transdeeplab: Convolution-free transformer-based deeplab v3+ for medical image segmentation,'' in \emph{International Workshop on PRedictive Intelligence In MEdicine}.\hskip 1em plus 0.5em minus 0.4em\relax Springer, 2022, pp. 91--102.

\bibitem{salman2023orthonets}
H.~Salman, C.~Parks, M.~Swan, and J.~Gauch, ``Orthonets: Orthogonal channel attention networks,'' \emph{arXiv preprint arXiv:2311.03071}, 2023.

\bibitem{shen2018disan}
T.~Shen, T.~Zhou, G.~Long, J.~Jiang, S.~Pan, and C.~Zhang, ``Disan: Directional self-attention network for rnn/cnn-free language understanding,'' in \emph{Proceedings of the AAAI conference on artificial intelligence}, vol.~32, no.~1, 2018.

\bibitem{chen2021transunet}
J.~Chen, Y.~Lu, Q.~Yu, X.~Luo, E.~Adeli, Y.~Wang, L.~Lu, A.~L. Yuille, and Y.~Zhou, ``Transunet: Transformers make strong encoders for medical image segmentation,'' \emph{arXiv preprint arXiv:2102.04306}, 2021.

\bibitem{cao2022swin}
H.~Cao, Y.~Wang, J.~Chen, D.~Jiang, X.~Zhang, Q.~Tian, and M.~Wang, ``Swin-unet: Unet-like pure transformer for medical image segmentation,'' in \emph{European conference on computer vision}.\hskip 1em plus 0.5em minus 0.4em\relax Springer, 2022, pp. 205--218.

\bibitem{resunet+}
S.~Metlek and H.~Çetıner, ``Resunet+: A new convolutional and attention block-based approach for brain tumor segmentation,'' \emph{IEEE Access}, vol.~11, pp. 69\,884--69\,902, 2023.

\bibitem{tscunet}
Z.~Jiang, C.~Ding, M.~Liu, and D.~Tao, ``Two-stage cascaded u-net: 1st place solution to brats challenge 2019 segmentation task,'' in \emph{Brainlesion: Glioma, Multiple Sclerosis, Stroke and Traumatic Brain Injuries}, A.~Crimi and S.~Bakas, Eds.\hskip 1em plus 0.5em minus 0.4em\relax Cham: Springer International Publishing, 2020, pp. 231--241.

\bibitem{hatamizadeh2022unetr}
A.~Hatamizadeh, Y.~Tang, V.~Nath, D.~Yang, A.~Myronenko, B.~Landman, H.~R. Roth, and D.~Xu, ``Unetr: Transformers for 3d medical image segmentation,'' in \emph{Proceedings of the IEEE/CVF winter conference on applications of computer vision}, 2022, pp. 574--584.

\bibitem{hatamizadeh2022swin}
A.~Hatamizadeh, V.~Nath, Y.~Tang, D.~Yang, H.~R. Roth, and D.~Xu, ``Swin unetr: Swin transformers for semantic segmentation of brain tumors in mri images,'' in \emph{International MICCAI Brainlesion Workshop}.\hskip 1em plus 0.5em minus 0.4em\relax Springer, 2021, pp. 272--284.

\bibitem{siddiquee2021redundancy}
M.~M. Rahman~Siddiquee and A.~Myronenko, ``Redundancy reduction in semantic segmentation of 3d brain tumor mris,'' in \emph{International MICCAI Brainlesion Workshop}.\hskip 1em plus 0.5em minus 0.4em\relax Springer, 2021, pp. 163--172.

\bibitem{futrega2021optimized}
M.~Futrega, A.~Milesi, M.~Marcinkiewicz, and P.~Ribalta, ``Optimized u-net for brain tumor segmentation,'' in \emph{International MICCAI brainlesion workshop}.\hskip 1em plus 0.5em minus 0.4em\relax Springer, 2021, pp. 15--29.

\bibitem{couplingunet}
K.~Kotowski, S.~Adamski, B.~Machura, L.~Zarudzki, and J.~Nalepa, ``Coupling nnu-nets with expert knowledge for accurate brain tumor segmentation from mri,'' in \emph{Brainlesion: Glioma, Multiple Sclerosis, Stroke and Traumatic Brain Injuries}, A.~Crimi and S.~Bakas, Eds.\hskip 1em plus 0.5em minus 0.4em\relax Cham: Springer International Publishing, 2022, pp. 197--209.

\end{thebibliography}

\end{document}